\documentclass[lettersize,journal]{IEEEtran}
\usepackage{amsmath,amsfonts}
\usepackage{array}
\usepackage[caption=false,font=normalsize,labelfont=sf,textfont=sf]{subfig}
\usepackage{textcomp}
\usepackage{stfloats}
\usepackage{url}
\usepackage{verbatim}
\usepackage{graphicx}
\usepackage{ragged2e}

\usepackage{cite}

\usepackage{physics}

\usepackage{float}
\usepackage{makecell}
\usepackage{multirow}      
\usepackage[table]{xcolor}    
\usepackage{tabularx}         
\usepackage{makecell}         
\usepackage{array}            
\usepackage{hhline}           
\usepackage{booktabs}         

\usepackage{makecell}      
\usepackage[table]{xcolor} 
\usepackage{needspace}                           
\usepackage{booktabs}  
\usepackage{multirow}  
\usepackage[table]{xcolor}
\usepackage{caption}
\usepackage{array}          
\usepackage{booktabs}       
\usepackage{multirow}       
\usepackage{pifont}
\usepackage{pifont}       

\usepackage[hidelinks]{hyperref}
\newcommand{\supcite}[1]{\textsuperscript{\cite{#1}}}

\begin{document}

\title{FUSAR-R1: A Large-Scale Reasoning Model for Intelligent Interpretation of SAR Images}

\author{Yi Yang, Xiaokun Zhang, Yuxuan Li, Ruyi Zhang, Xinpeng Zhou, Haipeng Wang
\thanks{ \textit{}
		
Yi Yang, Xiaokun Zhang, Yuxuan Li, Ruyi Zhang, Xinpeng Zhou, and Haipeng Wang are with the Key Laboratory for Information Science of Electromagnetic Waves (MoE), Fudan University, Shanghai 200433, China. \textit{(Corresponding author: Haipeng Wang.)}  

	}}

\markboth{Journal of \LaTeX\ Class Files,~Vol.~14, No.~8, August~2021}%
{Shell \MakeLowercase{\textit{et al.}}: A Sample Article Using IEEEtran. cls for IEEE Journals}


\maketitle

\begin{abstract}
In recent years, large-scale vision-language models have been driving a paradigm shift in intelligent remote sensing image interpretation. By incorporating textual semantic information, the cognitive expression, semantic understanding, and human-computer interaction capabilities of interpretation models have been significantly improved, achieving initial progress in the field of Synthetic Aperture Radar (SAR) image interpretation~\supcite{bao2005radar,xu2024microwavevision}. However, SAR images are affected by factors such as coherent imaging mechanisms, complex scattering characteristics, speckle noise interference, and target-background coupling, resulting in complex and variable image features with significant uncertainties and specializations. Existing SAR vision-language models do not yet possess the step-by-step analysis, logical judgment, and self-correction capabilities of human experts, making it difficult to support reliable intelligent interpretation in complex scenarios.
To address this issue, this paper proposes a large-scale reasoning model, FUSAR-R1, for intelligent interpretation of SAR images. The model first constructs explicit chain-of-thought(coT) reasoning data by simulating the interpretation process of human experts and uses this data to guide instruction learning, thereby endowing the model with basic reasoning capabilities. Subsequently, a reinforcement learning strategy is introduced to optimize the model’s outputs based on inference results, enabling self-correction and more reliable reasoning. Experimental results demonstrate that FUSAR-R1 consistently outperforms existing multimodal large-scale models across various SAR interpretation tasks, including target detection, target counting and classification, and land-cover category recognition.
\end{abstract}

\begin{IEEEkeywords}
SAR, vision-language model, chain-of-thought, reinforcement learning, artificial intelligence, foundation model
\end{IEEEkeywords}

\section{Introduction}
\IEEEPARstart{S}{ynthetic} Aperture Radar (SAR), as an active microwave remote sensing imaging technology, has all-weather, all-day and long-distance observation capabilities. It can stably acquire surface information under complex conditions such as clouds, fog, rain, snow, and night, and has been widely used in important fields such as military reconnaissance, marine monitoring, disaster assessment, resource surveys, and urban management. With the rapid development of high-resolution, multi-polarization, and multi-mode SAR sensors, the volume of SAR image data continues to increase, while observation scenarios are becoming increasingly complex, imposing higher demands on automated, intelligent, and reliable interpretation techniques. Traditional SAR image interpretation methods mainly rely on human expertise, physical models, or manually designed features~\supcite{jiao2008sar}, which can perform well in specific scenarios. However, when applied to large-scale datasets with multiple categories and complex backgrounds, these methods often suffer from limited generalization, a heavy reliance on human intervention, and limited cross-scenario adaptability.

To overcome these limitations, deep learning methods have been widely introduced into the intelligent interpretation of SAR images, significantly improving the performance of tasks such as target detection, target classification, ground feature segmentation, and scene recognition~\supcite{huang2017transfer}. However, SAR images differ from optical remote sensing images; their imaging results are determined by the complex scattering interactions between electromagnetic waves and ground features, commonly exhibiting phenomena such as speckle noise, geometric distortion, strong discrete scattering distribution, shadow occlusion, and target-background coupling~\supcite{singh2016analysis,baselice2009layover}. These factors lead to unstable target appearances and sparse semantic information in SAR images, thereby increasing the difficulty of feature extraction, target identification, and scene understanding. Especially in scenes with weak targets, dense targets, and complex backgrounds, models relying solely on end-to-end visual discrimination are insufficient to fully explain the scattering characteristics, spatial structure, and geographic semantic relationships of targets.

Therefore, the intelligent interpretation of SAR images is gradually evolving from the traditional stage of “perception and recognition” toward higher levels of “semantic cognition” and “logical reasoning”. Future SAR interpretation models should not only possess stable visual feature perception capabilities but also be able to comprehensively analyze imaging mechanisms, spatial scales, geographical contexts, and task semantics, thereby generating more accurate, interpretable, and reliable interpretation results.

In recent years, the development of large-scale vision–language models has provided a new technical pathway for intelligent interpretation of remote sensing images. By incorporating natural-language semantic information, these models no longer merely output category labels, bounding boxes, or segmentation maps but can instead generate textual descriptions of target attributes, scene relationships, and interpretation results, thereby significantly enhancing semantic expressiveness and human–computer interaction capabilities. In the field of SAR image interpretation, existing research has begun to explore cross-modal alignment between SAR imagery and textual semantics, enabling models to perform tasks such as image captioning, target counting, visual question answering, and spatial localization~\supcite{wei2025sarlang,ma2025sarclip,10642118}. However, most existing SAR vision–language models still primarily focus on image–text matching, semantic generation, or instruction following. Their outputs are typically directly oriented toward final answers, while lacking explicit modeling of scattering mechanisms, target structures, spatial relationships, and geographic context. Therefore, enabling models not only to “provide answers” but also to “explain their reasoning, perform step-by-step inference, and self-correct errors” has become a key challenge for the further development of intelligent SAR image interpretation.

Meanwhile, the field of artificial intelligence (AI) is also undergoing a transition from generative AI to reasoning-oriented AI. Large-scale reasoning models, such as OpenAI o1~\supcite{openai2024o1} and DeepSeek-R1~\supcite{guo2025deepseek}, demonstrate that by introducing long-chain reasoning, outcome feedback, and reinforcement learning-based optimization, models can decompose complex problems, perform step-by-step inference, and self-correct prior to producing final outputs, thereby improving logical consistency and result reliability in complex tasks. This paradigm is also beginning to influence research on intelligent remote sensing interpretation, where some reasoning-oriented models attempt to enhance spatial localization, target counting, and pointing-based representation understanding through explicit chain-of-thought supervision and reinforcement learning strategies~\supcite{liu2025visual}. However, existing research primarily focuses on optical remote sensing images, with reasoning processes revolving around spatial location, semantic relationships, and visual context. For SAR images, reliable interpretation must also incorporate expertise such as electromagnetic scattering mechanisms, speckle noise, shadow structure, target scale, and geographical priors. Therefore, constructing a large-scale reasoning model that integrates thought chain reasoning, reinforcement learning optimization, and the physical characteristics of SAR imaging is a key direction for achieving highly reliable intelligent SAR interpretation.

However, building a large-scale inference model for intelligent interpretation of SAR images still faces many challenges.

(1) The SAR image interpretation process has not yet been adequately modeled as a reliable and explicit reasoning procedure. Although existing SAR vision–language models can perform tasks such as image description, target counting, visual question answering, and spatial localization, their core capabilities are still mainly reflected in image–text semantic alignment and instruction following, typically generating final answers directly from input images. For SAR imagery, target interpretation is not merely a visual feature-matching process, but requires comprehensive analysis of multiple cues, including strong scattering highlights, shadow structures, edge responses, target scale, orientation relationships, and geographic context. Therefore, constructing a dedicated reasoning framework tailored to SAR imaging characteristics and target interpretation logic—enabling the model to perform explicit reasoning following the process of “observation cues → physical analysis → spatial verification → conclusion generation”—is the first key challenge that needs to be addressed.

(2) There is a lack of high-quality chain-of-thought data to support SAR reasoning learning. Most existing SAR image datasets only provide result-oriented annotations, such as categories, locations, quantities, or brief textual descriptions. Although these annotations can support models in learning the mapping relationship between images and answers, they are insufficient to capture the intermediate analytical process that experts follow when interpreting SAR imagery. Without structured chain-of-thought data, models tend to remain at a surface-level semantic matching stage, making it difficult to learn the expert interpretation paradigm of “which cues to follow and in what order to reach a conclusion”.

(3) There is a lack of feedback optimization and self-correction mechanisms for the SAR inference process. While thought chain supervision can guide the model to generate intermediate inference processes to some extent, it is essentially still a form of imitation learning of expert corpus distribution. In complex scenarios, the model may still suffer from reasoning hallucinations, logical redundancy, insufficient physical evidence, or erroneous conclusions. For SAR images, targets are often affected by factors such as speckle noise, sidelobe response, strong scattering background, occlusion, and target-background coupling. The model not only needs to determine whether the final answer is correct but also needs to ensure that the inference path conforms to electromagnetic scattering mechanisms, scale constraints, and spatial semantic relationships.

(4) There is a lack of a unified inference optimization mechanism for multi-task SAR interpretation. SAR intelligent interpretation tasks are not limited to single-target recognition, but instead involve multiple task forms, including target-level localization, object counting, category classification, scene semantic understanding, and region proportion estimation. When a single reward function is used for optimization, it is difficult to accommodate the differing objectives across tasks, and it also fails to effectively constrain the model to produce structurally consistent, semantically correct, and spatially coherent inference results in multi-task scenarios.

Inspired by the above research, this paper makes the following main contributions to the SAR image target interpretation task: 

(1) A large-scale inference model, FUSAR-R1, for intelligent SAR image interpretation is proposed. This model, with explicit inference at its core, extends the SAR image interpretation process from traditional end-to-end result prediction to an interpretable logical inference process. This allows the model to perform step-by-step analysis around target scattering characteristics, spatial structure, scale information, and geographical environment, thereby improving the reliability of interpretation results in complex scenarios.
(2) Constructing a SAR interpretation thought chain dataset: A structured thought chain data construction pipeline was designed to provide the model with a complete inference path from SAR images to target attributes (category, quantity, location).

(3) Design a reinforcement learning-driven reasoning optimization mechanism: Based on the cold start training of the thought chain, a reinforcement learning strategy is introduced to constrain the model's reasoning process through task feedback, guide the model to autonomously correct reasoning biases, and continuously optimize the quality of the thought chain and the credibility of the interpretation.

(4) This paper designs a unified reward function for multi-task SAR interpretation. For different task formats, including target detection, target counting and category recognition, land-cover classification, and region understanding, a multi-task reward mechanism is constructed from four aspects: format reward, target detection reward, target counting and category-matching reward, and land-cover classification reward. This enables the model to be simultaneously constrained by multi-dimensional feedback signals, including output structure, spatial localization, target attributes, and scene semantics during reinforcement learning, thereby improving the inference stability and result reliability of FUSAR-R1 in multi-task SAR interpretation scenarios.

\section{Related work}
\subsection{Research on vision-language models in the field of remote sensing}
In recent years, vision–language models have achieved significant progress in natural image understanding. By jointly modeling visual and textual modalities, these models map visual features into a semantic language space, providing a new paradigm for tasks such as open-vocabulary recognition, image captioning, visual question answering, and human–computer interaction. Inspired by these advances, the remote sensing community has also begun to explore cross-modal semantic alignment, visual instruction understanding, and open-world scene interpretation. Unlike traditional remote sensing models that primarily output category labels, bounding boxes, or segmentation masks, vision–language models for remote sensing can leverage natural language to describe land-cover categories, spatial relationships, scene functions, and regional attributes, thereby significantly enhancing semantic expressiveness and adaptability to open-set tasks.

In terms of image-text alignment, RemoteCLIP~\supcite{10504785} borrows the contrastive learning approach from Contrastive Language--Image Pretraining (CLIP), using category labels from publicly available remote sensing datasets to construct templated text descriptions. This transforms the original category supervision into image-text alignment signals, thereby enhancing the open category transfer capability of remote sensing visual representations. SkyScript~\supcite{wang2024skyscript} further utilizes geographic semantic information from OpenStreetMap to automatically generate fine-grained text descriptions for large-scale optical remote sensing images, alleviating the problem of insufficient remote sensing image-text pairing data and improving the scale and semantic richness of image-text alignment data. Addressing the potential semantic illusions that vision-language models may produce in complex remote sensing scenarios, VHM~\supcite{pang2025vhm} leverages the logical organization capabilities of large language models to construct a high-quality HqDC dataset. Through more rigorous data filtering and semantic association, it improves the realism and consistency of image-text descriptions.

In terms of command understanding and interactive interpretation, SkyEye-GPT~\supcite{zhan2025skyeyegpt} and GeoChat~\supcite{Kuckreja_2024_CVPR}, by constructing visual command data and multi-turn dialogue samples, enable remote sensing vision-language models to perform scene question answering, target localization, and region description in response to natural language commands. SkyEye-GPT employs a two-stage fine-tuning strategy to improve the model's command following ability and dialogue logic, while GeoChat emphasizes open-ended question answering and geographic semantic interaction in remote sensing scenes. These methods are driving remote sensing models from traditional closed visual recognition to more flexible human-computer collaborative interpretation.

In addition, some research has begun to improve remote sensing vision-language modeling capabilities by focusing on model structure and task adaptation. BITA~\supcite{10415446} introduces a lightweight Fourier Transformer structure to enhance the depth of interaction between image and text features through frequency domain modeling, allowing the model to better capture texture, scale, and spatial distribution information in remote sensing images. BAN~\supcite{10438490} builds a special basic model for remote sensing change detection tasks, which strengthens the model's ability to perceive the spatiotemporal evolution characteristics of ground objects.
Overall, existing remote sensing vision–language models have evolved from early image–text alignment to stages of instruction-based interaction, region-level understanding, and task-specific modeling. However, most of these methods primarily focus on optical remote sensing imagery, where model understanding relies heavily on color, texture, and appearance-based semantics. As a result, they are difficult to directly adapt to the more abstract visual representations in SAR imagery, which are jointly determined by electromagnetic scattering mechanisms, shadow structures, and imaging geometry.

\subsection{Research on vision-language models in the field of SAR}
Compared with the rapid development of vision–language models in optical remote sensing, multimodal semantic modeling for SAR imagery is still in its early stages. SAR images are acquired through active microwave imaging, and their visual representations are primarily determined by electromagnetic scattering intensity, target structure, material properties, incidence angle, shadow effects, and background conditions, lacking the intuitive color and texture semantics present in optical imagery. As a result, a more pronounced semantic gap exists between SAR images and natural language. On the one hand, the same target may exhibit significantly different scattering responses under varying imaging conditions and observation angles. On the other hand, the availability of SAR image–text paired data remains limited, and text annotations typically rely on specialized imaging knowledge and expert interpretation. Consequently, training vision–language models for SAR imagery is considerably more challenging than for optical remote sensing scenarios.

Early research on SAR vision–language modeling primarily attempted to establish a preliminary association between SAR imagery and category semantics. Works such as SARLANG~\supcite{wei2025sarlang} and SAR-CLIP~\supcite{ma2025sarclip} leveraged existing detection or classification annotations to generate textual descriptions, converting structured labels such as target categories and scene attributes into language supervision signals, thereby achieving an initial alignment between SAR images and textual semantic spaces. Although these methods demonstrated the feasibility of cross-modal representation learning for SAR imagery, their textual supervision was typically derived from reverse descriptions of category labels or detection outputs. As a result, the semantic richness of the generated text remains limited, making it difficult to fully capture the scattering mechanisms, target structures, scale constraints, and complex spatial relationships inherent in SAR imagery.

To reduce the cost of SAR image annotation, SSL-LIP~\supcite{10642118} proposes a SAR image-text alignment model under limited annotation constraints. This model utilizes a two-stage self-supervised training paradigm to mine the potential correlations between SAR images and language, providing an important approach for SAR multimodal representation learning under low annotation cost conditions. While this type of method alleviates the problem of insufficient SAR text resources to some extent, its main goal remains improving image-text matching and semantic alignment capabilities, and it has not yet further addressed the deep mapping problem between physical scattering cues in SAR images and linguistic knowledge.

Building upon this foundation, the Fudan University team proposed FUSAR-KLIP~\supcite{yang2025fusarklip} and FUSAR-GPT~\supcite{Zhang_2026_CVPR}, further advancing the development of SAR vision-language models. FUSAR-KLIP addresses the cognitive gap between SAR images and human semantic space by constructing the large-scale SAR image-text dataset FUSAR-GEOVL-1M. Through knowledge-guided text generation and cross-modal alignment strategies, it organizes target categories, scene attributes, imaging features, and geographic semantic information into a unified visual language representation space, thereby enhancing the model's ability to understand the semantic content of SAR images. FUSAR-GPT further develops towards generative multimodal large-scale models by introducing multi-source remote sensing feature embedding and visual command fine-tuning, enhancing the model's interactive interpretation capabilities in tasks such as SAR image description, scene question answering, target counting, and spatial localization.

Overall, existing SAR vision-language models have evolved from initial category semantic alignment to a stage of large-scale model interaction involving knowledge-guided image-text representation learning and multi-source prior enhancement, laying the foundation for open-ended understanding of SAR images. However, most of these methods still primarily focus on image-text alignment, target description, question-and-answer generation, or command response. The models typically output the final answer directly, lacking explicit, multi-step analysis of target geometry, scattering response, spatial layout relationships, scale constraints, and geographic environmental information. In other words, while existing SAR vision-language models have improved the semantic understanding of SAR images, they have not yet truly possessed the step-by-step reasoning, logical verification, and self-correction capabilities of human experts, making it difficult to meet the demands of highly reliable intelligent SAR interpretation in complex scenarios.

\subsection{Current status of the development of reasoning models}
In recent years, the field of general artificial intelligence has been rapidly evolving from generative models to reasoning models. Traditional large language models mainly rely on supervised fine-tuning or instruction alignment to learn the mapping relationship between input and output. Although they can generate fluent answers, their reasoning process is usually implicit in the model parameters, lacking interpretable intermediate analysis paths. Reasoning models, represented by OpenAI o1 and DeepSeek-R1, demonstrate that by introducing long-chain thinking, result feedback, and reinforcement learning optimization, the model can decompose the problem, deduce step by step, self-check, and correct errors before giving the final answer, thus significantly improving the logical consistency and result credibility in complex tasks. Inspired by this, related research such as Visual-RFT~\supcite{liu2025visual} further extends the R1 paradigm to vision-language models, enhancing the model's reasoning ability in visual question answering, object counting, spatial localization, and complex scene understanding through supervised fine-tuning of thought chains, reinforcement learning fine-tuning, and task-related reward design.

In the field of remote sensing, research on inference models has also begun to attract attention. Due to the characteristics of remote sensing imagery, such as large observation scale, dense targets, complex spatial relationships, and strong semantic ambiguity, many interpretation tasks cannot be completed simply by class recognition. Instead, models need to comprehensively judge based on target location, proximity relationships, land cover functions, and contextual environment. For example, RSThinker~\supcite{liu2025towards}, by constructing the Geo-CoT thought chain dataset, models the remote sensing analysis process as a verifiable inference chain of task planning, evidence localization, and conclusion synthesis, and combines supervised fine-tuning and Group Relative Policy Optimization (GRPO) reinforcement learning to improve the authenticity of the inference results. Geo-R1~\supcite{zhang2025geo}, targeting small-sample geospatial indexing interpretation tasks, introduces reinforcement learning fine-tuning, enabling the model to generate explicit inference chains and complete localization based on target-context relationships. RSGround-R1~\supcite{huang2026rsground} further emphasizes the spatial reasoning problem in remote sensing visual localization, improving the model's localization stability through location-aware thought chains, distance-sensitive reward mechanisms, and spatial consistency optimization.

Overall, existing remote sensing inference models primarily revolve around optical remote sensing images. Their inference processes focus more on spatial location, semantic relationships, and visual context, while neglecting the unique electromagnetic scattering mechanisms, speckle noise, shadow structure, scale constraints, and imaging geometry of SAR images. Therefore, constructing large-scale inference models with physical consistency, spatial reasoning capabilities, and self-correction abilities for intelligent interpretation of SAR images remains a crucial direction that urgently needs breakthroughs in current SAR visual language research.

\section{Methods}
To address the aforementioned issues, this paper proposes a large-scale inference model, FUSAR-R1, for intelligent interpretation of SAR images. This model explicitly models the SAR interpretation process as a chain of thoughts and introduces Group Relative Policy Optimization (GRPO)~\supcite{shao2024deepseekmath} as a reinforcement learning training paradigm.

\subsection{Overall Framework}
\begin{figure}
    \centering
    \includegraphics[width=1.1\linewidth]{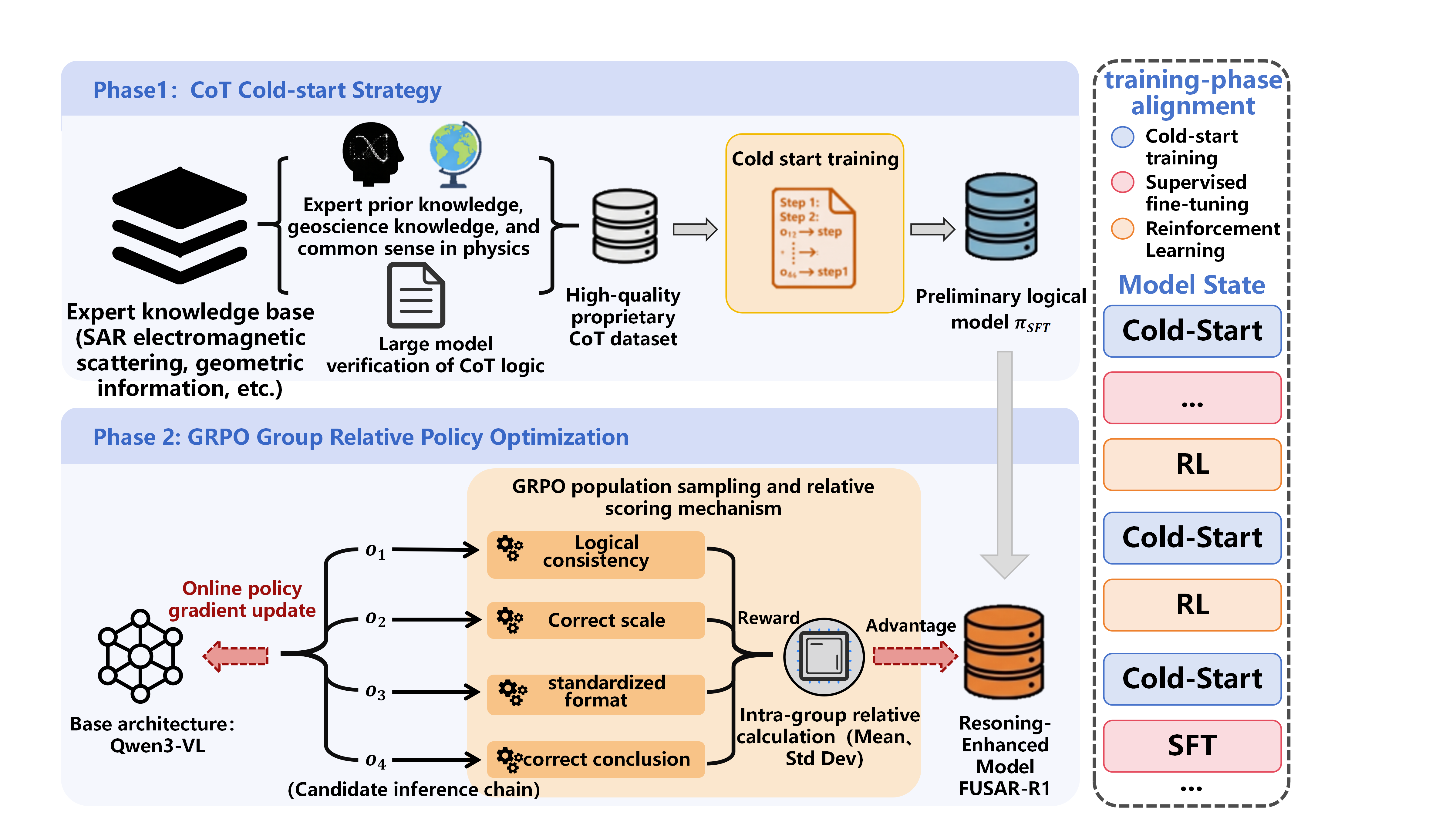}
    \caption{overall framework of FUSAR-R1}
    \label{fig：fig01}
\end{figure}

This paper proposes the FUSAR-R1 inference model, the overall architecture of which is shown in Figure \ref{fig：fig01}. This model uses the newly released open-source multimodal large model Qwen3-VL as its underlying backbone architecture, fully leveraging its superior performance in multi-scale visual perception and complex instruction following.

In terms of training strategy, this study first reconstructs the FUSAR-GEOVL-1M dataset through deep logical reconstruction, transforming originally fragmented semantic labels into structured textual descriptions with rigorous deductive logic, thereby constructing a high-quality, expert-level chain-of-thought cold-start corpus.

Subsequently, to address the specialized requirements of the SAR image interpretation tasks, a GRPO-based reinforcement learning strategy is designed. By constructing a multi-task reward function composed of sub-rewards, including format constraints, object detection, object counting, category recognition, and land-cover classification, the model is guided to optimize inference quality across different SAR interpretation tasks while maintaining output structural consistency.

\subsection{SAR Interpretation Chain-of-Thought Data Construction}
This paper reconstructs the thought process based on the previously constructed FUSAR-GEOVL-1M dataset. FUSAR-GEOVL-1M is a large-scale multimodal dataset for vision-language modeling of SAR images. It preserves the textual semantics, target categories, scene attributes, and geospatial information corresponding to SAR images, providing a data foundation for cross-modal alignment between SAR images and natural language. However, the original FUSAR-GEOVL-1M mainly provides descriptive labels corresponding to images. Although it contains some target categories, scene attributes, and geographic semantic information, its textual form is more inclined towards the final conclusion and lacks the reasoning process between image observation and interpretation results. To compensate for this deficiency, this paper utilizes the locally deployed Qwen3-8B large model as the logic reconstruction engine and designs a rigorous expert interpretation pipeline to transform the originally fragmented semantic information into a structured reasoning path with electromagnetic scattering logic, spatial scale constraints, and uncertainty assessment.

Specifically, in the pipeline design, metadata information of the image, including satellite type, imaging band, ground resolution, and latitude–longitude coordinates, is first used as input. This information serves as the initial triggering condition for generating the inference chain, guiding the model to perform geographic anchoring of the scene and to infer potential target categories and their spatial distribution patterns based on the surrounding geographic context. For example, when the latitude–longitude coordinates correspond to a specific airport or port, the model is guided by contextual cues to leverage its geographic priors and commonsense knowledge embedded in its parameters, thereby predicting the types of targets that are likely to appear in the region and their corresponding spatial distribution patterns.

Subsequently, this paper explicitly introduces physical verification logic and specific SAR scattering feature descriptions during the thought chain generation process. Targets in SAR images typically exhibit features such as bright spots, shadows, edge enhancement, double-bounce scattering, wakes, or combinations of strong and weak scattering. The imaging performance of different targets is closely related to their structural morphology, material properties, attitude orientation, and background environment. This paper constructs a comprehensive " common scattering feature library" , stipulating that the model must retrieve and compare these visual phenomena during inference. Taking aircraft targets as an example, the model no longer classifies them solely based on outlines but is required to infer target information based on specific structural features such as wings and engine nacelles. Therefore, during the thought chain generation process, the model not only needs to describe "the existence of a certain type of target in the image" but also needs to explain which observable cues this judgment comes from.

Furthermore, scale prior is also a crucial link in the thought process. The model must use image resolution for quantitative calculations to establish a mapping relationship between pixel span and real size: for example, a bright spot with a length of 200 pixels corresponds to a physical length of approximately 100 meters at a resolution of 0.5 meters, thus helping to distinguish similar aircraft models such as the Airbus A330 and Airbus A350. By incorporating spatial scale, physical parameters, and scattering characteristics into the reasoning process, the model can move from simple feature matching to more physically based logical interpretation.

For different interpretation tasks, this paper further designs differentiated interpretation logics: aircraft target interpretation focuses on the airframe structure and its spatial relationship with background elements such as runways, taxiways and hangars; ship target interpretation emphasizes the target contrast, outline features and wake information against the background of the sea surface; and the land feature segmentation task focuses on regional consistency, functional zone division and the constraints of the geographical environment on the distribution of cities, waterways and transportation corridors.

\begin{figure}
    \centering
    \includegraphics[width=1\linewidth]{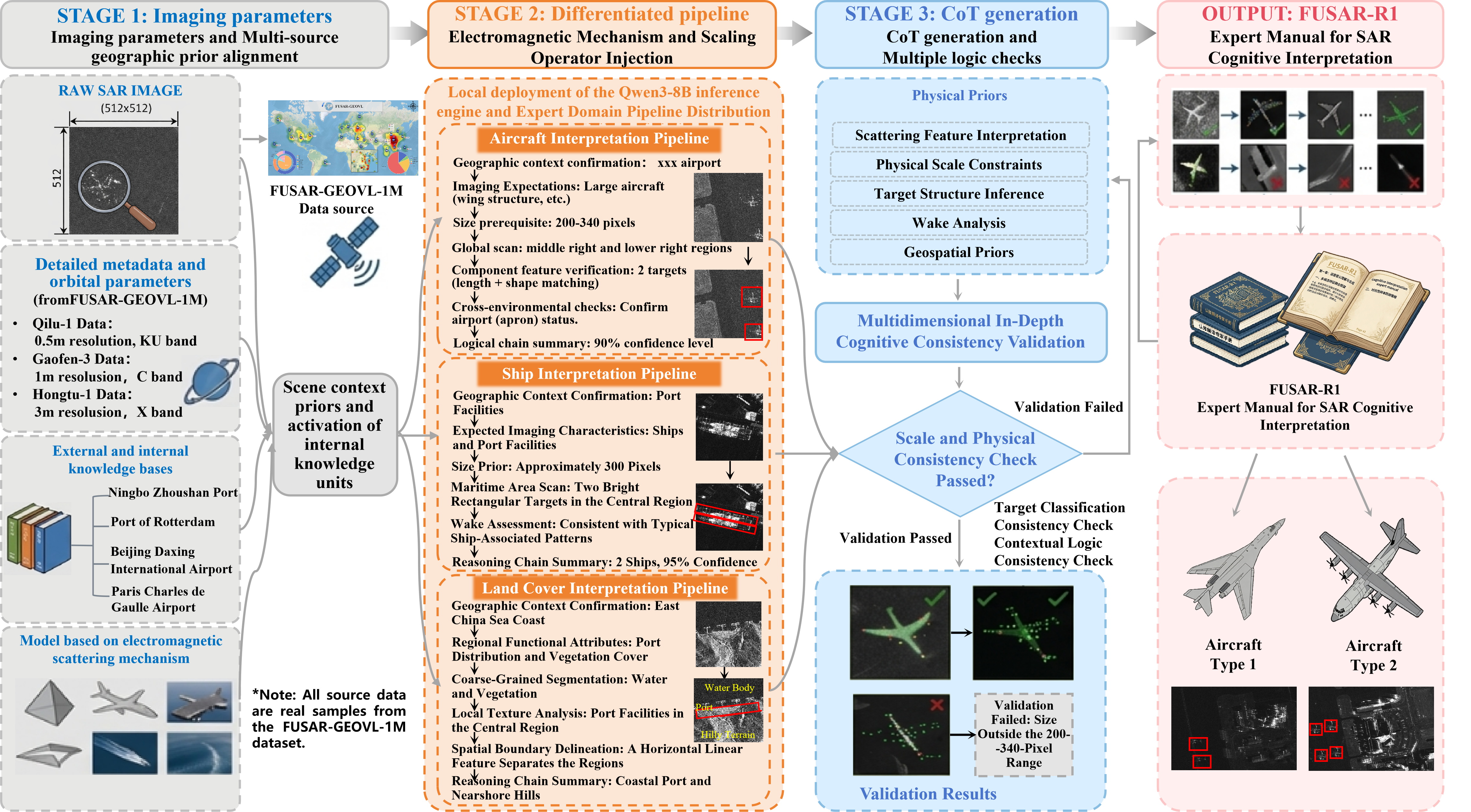}
    \caption{SAR interpretation mind chain data construction pipeline}
    \label{fig：fig02}
\end{figure}

This categorized and customized pipeline ensures that the FUSAR-R1 corpus maintains expert-level logical rigor and physical consistency while covering millions of samples.

In the final stage of dataset reconstruction, this paper introduces an adaptive correction and logical consistency filtering mechanism. On the one hand, because the textual descriptions in FUSAR-GEOVL-1M may be incomplete, the Qwen3-8B model is encouraged to leverage its inherent cross-domain knowledge to supplement and correct the original descriptions during CoT generation, thereby producing more complete analytical paths. On the other hand, samples with obvious errors, such as unreasonable scale estimation, conflicts between target categories and geographic environments, or scattering feature interpretations that violate physical mechanisms, are automatically filtered or regenerated.

\newcolumntype{L}{>{\raggedright\arraybackslash}X}

\newcolumntype{L}{>{\centering\arraybackslash}X}
\newcolumntype{s}{>{\centering\arraybackslash}p{2cm}}

\begin{table*}[htbp]
	\centering
	\caption{Data status of FUSAR-R1 and FUSAR-GEOVL-1M}
	\label{tab:tab01_cot_comparison}
	\small 
	\begin{tabularx}{\textwidth}{@{} l s s L @{}}
		\toprule
		\textbf{Interpretation Task} & \textbf{FUSAR-GEOVL-1M} & \multicolumn{2}{c}{\textbf{FUSAR-R1 (Ours)}} \\ 
		\cmidrule(lr){2-2} \cmidrule(lr){3-4}
		& \textbf{Reasoning Steps} & \textbf{Reasoning Steps} & \textbf{Reasoning Process Decomposition} \\ \midrule
		
		\textbf{Aircraft} & \makecell{1 Step \\[2pt] (Description Only)} & \textbf{7 Steps} & 
		1. Geographic Context Confirmation $\rightarrow$ 2. Aircraft Imaging Expectation $\rightarrow$ 3. Scale Prior $\rightarrow$ 4. Global Scanning $\rightarrow$ 5. Aircraft Component Feature Verification $\rightarrow$ 6. Environmental Cross-checking $\rightarrow$ 7. Logical Chain Summary \\ \addlinespace
		
		\textbf{Ship} & \makecell{1 Step \\[2pt] (Description Only)} & \textbf{6 Steps} & 
		1. Geographic Context Confirmation $\rightarrow$ 2. Ship Imaging Expectation $\rightarrow$ 3. Scale Prior $\rightarrow$ 4. Sea Area Scanning $\rightarrow$ 5. Wake Detection $\rightarrow$ 6. Logical Chain Summary \\ \addlinespace
		
		\textbf{Land-cover Classification} & \makecell{1 Step \\[2pt] (Description Only)} & \textbf{6 Steps} & 
		1. Geographic Context Confirmation $\rightarrow$ 2. Regional Functional Attributes $\rightarrow$ 3. Coarse-grained Structured Segmentation $\rightarrow$ 4. Local Texture Analysis $\rightarrow$ 5. Spatial Boundary Delineation $\rightarrow$ 6. Logical Chain Summary \\ \bottomrule
		
	\end{tabularx}
	\begin{flushleft}
		\footnotesize Note: FUSAR-GEOVL-1M only provides a single descriptive label (such as "There are three aircraft on the tarmac") and lacks intermediate reasoning.
	\end{flushleft}
\end{table*}

Finally, this article organizes each sample into a unified format of "image input - metadata description - thinking chain reasoning - final answer". Among them, the thinking chain part is used to explicitly present the intermediate process from SAR image observation to target attribute judgment, and the final answer corresponds to the specific interpretation result, such as target category, target number, spatial location, or surface object type. Compared with the original FUSAR-GEOVL-1M data set, which only provides conclusive descriptions, the data set constructed in this article breaks down the interpretation task into multiple interrelated logical links. This multi-step derivation model provides a rigorous physical basis for interpretation and can effectively reduce misjudgments when the model encounters complex SAR backgrounds.

\subsection{Startup training based on thought chain instructions}
After constructing a complete chain-of-thought dataset for SAR interpretation, this section introduces the key stage of model training, namely the cold-start process. The core objective of this stage is not to enable the model to explore from scratch through large-scale reinforcement learning, but to inject logical reasoning capabilities into the model via chain-of-thought-guided cold start training. The cold-start process not only equips the model with basic interpretation capabilities but, more importantly, establishes the linguistic style and logical norms that the model should follow when processing SAR imagery. This helps prevent logical collapse caused by an excessively large search space during the early stage of reinforcement learning.

We define the cold start process as a conditional probability maximization task on the expert corpus distribution $D_{expert}$. Given an input FUSAR-GEOVL-1M image $I$ and associated metadata $M$ (including resolution, band, latitude and longitude, etc.), the model needs to predict the complete sequence containing the thought chain $T$ and the interpretation conclusion $A$. We formulate the cold-start stage as an autoregressive supervised fine-tuning (SFT) process over the expert corpus $D_{\mathrm{expert}}$:
\begin{equation}
\mathcal{L}_{SFT}(\theta) = -\mathbb{E}_{(I, M, T, A) \sim D_{expert}} \left[ \sum_{t=1}^{|T|+|A|} \log P_{\theta}(y_t | I, M, y_{<t}) \right]
\end{equation}
Here, $\theta$ represents the model parameters, and $y_t$ represents the $t$th token in the inference sequence. This objective function is essentially an autoregressive supervised fine-tuning process, but unlike traditional instruction-based fine-tuning that only supervises the final answer $A$, the cold start phase forces the model to model the intermediate inference path $T$. This design, through maximum likelihood estimation, transforms the physical information defined in the thought chain, such as imaging resolution, geographic information, and bands, into the model's autoregressive generation strategy through parameterized learning. This allows the model to not only learn "what result to output" but also "what clues and in what order to obtain that result".

During the cold-start process, explicit constraints on the reasoning paradigm are crucial for enabling the model to acquire stable inference capabilities. SAR image interpretation is not a simple visual–semantic matching process; rather, it requires step-by-step judgment based on imaging conditions, scattering responses, target scale, and geographic context. Therefore, this paper requires the model to organize the intermediate analysis process using specific reasoning tags when generating the inference sequence and to provide the interpretation conclusion within the final answer tag. Through this format constraint, the model output no longer jumps directly from the image to the answer, but is instead guided into a continuous reasoning process consisting of “observational cue extraction → physical evidence analysis → spatial relationship verification → final conclusion generation”.

Furthermore, to ensure that cold-start corpora possess not only linguistic logical coherence but also the physical consistency required for SAR interpretation, this paper constructs inference paradigms with physical hard constraints for different target domains:

For aeronautical targets, the inference chain first uses latitude and longitude information to confirm the geographic context, determining whether the image belongs to an airport area, and then combines airport functional attributes and resident aircraft type information to form a set of candidate targets. Subsequently, the model predicts the strong scattering structure that the aircraft may present based on SAR imaging bands and resolution, and further uses scale priors for quantitative verification. Only when a candidate target simultaneously meets the requirements of length range, wing lateral extension characteristics, and spatial relationship with airport elements such as runways, taxiways, and aprons will the inference chain identify it as the corresponding aircraft type.

For ship targets, the cold-start corpus emphasizes the joint constraints among the sea background, target scale, and scattering mechanisms. In port scenarios, the model focuses on distinguishing dihedral scattering from background contrast. Specifically, the model learns that near-specular reflection from the water surface usually produces dark regions, whereas the straight decks and sidewalls of ships often generate extremely strong echoes. For example, “A high-brightness rectangular region is observed near the central berth, with a pixel span of approximately 100 pixels. Given a ground resolution of 0.5 m, this corresponds to a physical length of about 50 m, which is highly consistent with the scale characteristics of civilian ships. In addition, its vertical arrangement relative to the dock topology helps eliminate potential false alarms, supporting its identification as a berthed civilian ship”.

For land cover interpretation tasks, cold-start corpora focus on regional consistency and functional attribute inference. Unlike discrete target recognition, land cover interpretation requires models to start from the overall spatial layout and combine texture roughness, boundary continuity, regional morphology, and geographic environment to determine the type of land cover. When processing river network plain images such as those of Jiujiang, the model is required to deduce land cover attributes through texture roughness and regularity: "The upper part of the image shows densely arranged regular bright block textures, which are identified as urban building clusters. The wide linear feature spanning the middle has sharp edges, which are inferred to be a major transportation corridor or embankment. The large dark area in the lower part is consistent with the low reflectivity characteristics of water bodies".

The model weights $W_{cold}$ obtained during the cold-start phase provide a stable initial distribution for subsequent reinforcement learning optimization. Through training at this stage, the model learns to organize language in a structured manner and connect fragmented electromagnetic scattering features into a rigorous logical chain. This implantation of “logical seeds” substantially alleviates the reward sparsity problem during reinforcement learning, enabling the model to more efficiently explore higher-level SAR cognitive reasoning behaviors under the guidance of physically consistent reward functions.

\subsection{Optimization of Model Inference Based on Reinforcement Learning}

After completing cold-start training guided by expert chain-of-thought data, FUSAR-R1 initially acquires SAR image interpretation capabilities and can generate reasoning texts containing both intermediate analytical processes and final conclusions in a specific format. However, its generation process still essentially relies on maximum likelihood estimation under the training data distribution, making it susceptible to reasoning hallucinations, logical redundancy, and insufficient robustness in complex interference scenarios. To further enhance the model’s autonomous exploration capability and reasoning reliability, this paper introduces a reinforcement learning optimization strategy based on the cold-start model and adopts Group Relative Policy Optimization (GRPO) as the core training paradigm.

\subsubsection{Reinforcement learning}

We treat FUSAR-R1 as an agent and the interpretation scenario, comprised of SAR imagery, metadata, physical rules, and geographical prior knowledge, as the environment. During the autoregressive generation process, the model's state $s_t$ at time $t$ consists of the input image features, metadata information, and the currently generated sequence of inference tokens; the action $a_t$ corresponds to the next token predicted by the model. As tokens are generated, the model continuously completes the inference process from image observation, scattering feature analysis, spatial relationship judgment to the final answer output. The environment returns a reward $r_t$ and updates the state to $s_{t+1}$ based on whether the generated content meets format specifications, conforms to SAR imaging physics, correctly describes target attributes, and the accuracy of the final interpretation result. The core objective of reinforcement learning is to learn the optimal policy $\pi(a|s)$ that maximizes the expected value of the long-term cumulative discount reward.
\begin{equation}
	G_t = \sum_{k=0}^{\infty} \gamma^k r_{t+k}
\end{equation}
Here, $\gamma \in [0, 1]$ is a discount factor, which determines the model's emphasis on long-term interpretation accuracy.

This action-reward-based iterative optimization mechanism allows the model to actively explore a broad inference space, rather than simply relying on imitating expert templates. Through continuous state transitions and reward feedback, FUSAR-R1 can gradually internalize SAR interpretation logic; when a certain type of inference path can consistently obtain higher rewards, its generation strategy will be strengthened in gradient updates. Compared to purely supervised learning, reinforcement learning provides the model with an exploration space that surpasses manually labeled distributions through objective rule reward functions, enabling it to form simpler, more rigorous, and more generalizable inference paths, thereby improving the interpretation reliability in complex SAR scenarios.
\begin{figure}
    \centering
    \includegraphics[width=1\linewidth]{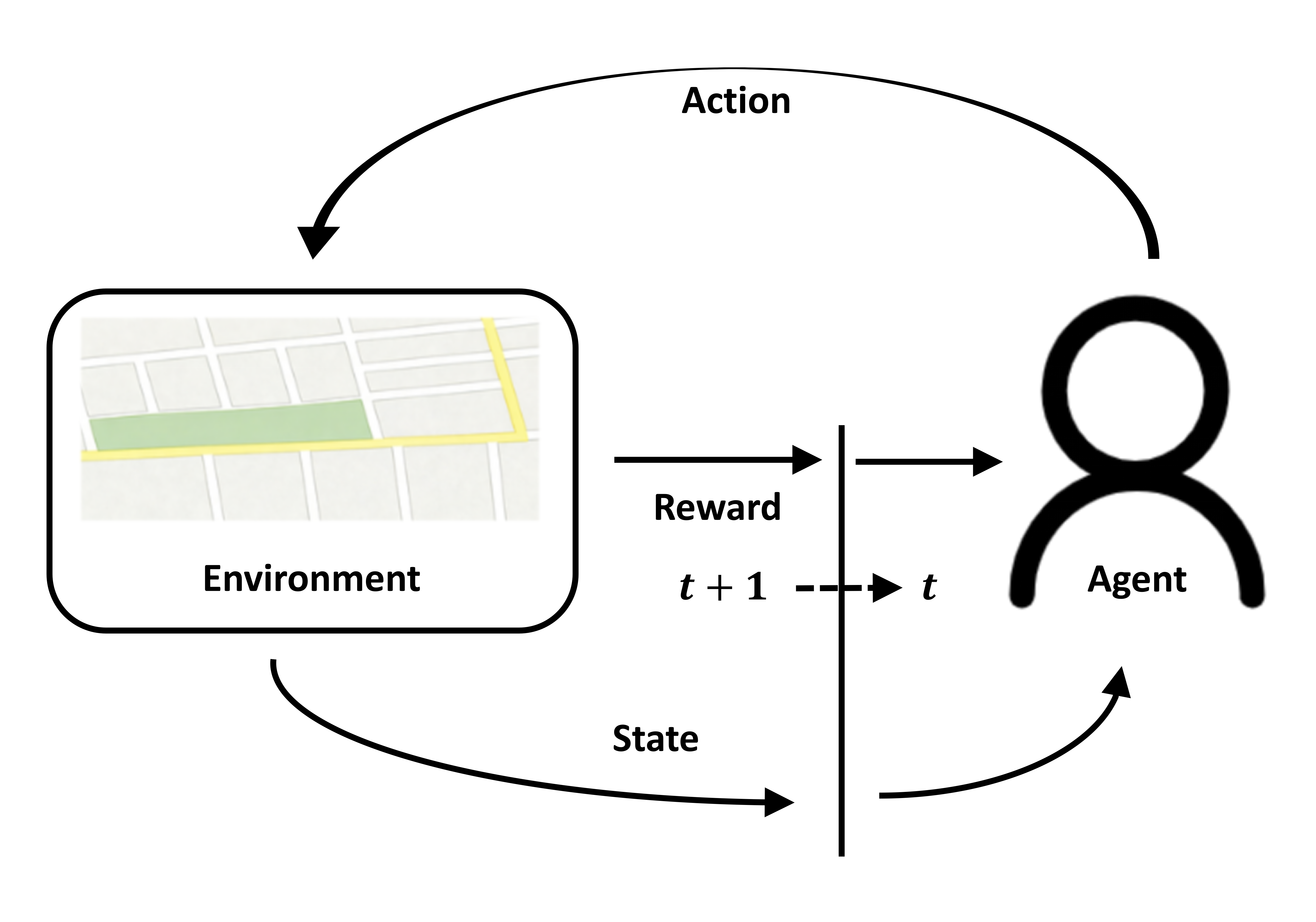}
    \caption{Basic process of reinforcement learning}
    \label{fig:fig03}
\end{figure}

\subsubsection{Group Relative Policy Optimization Algorithm} 

\begin{figure}
    \centering
    \includegraphics[width=1\linewidth]{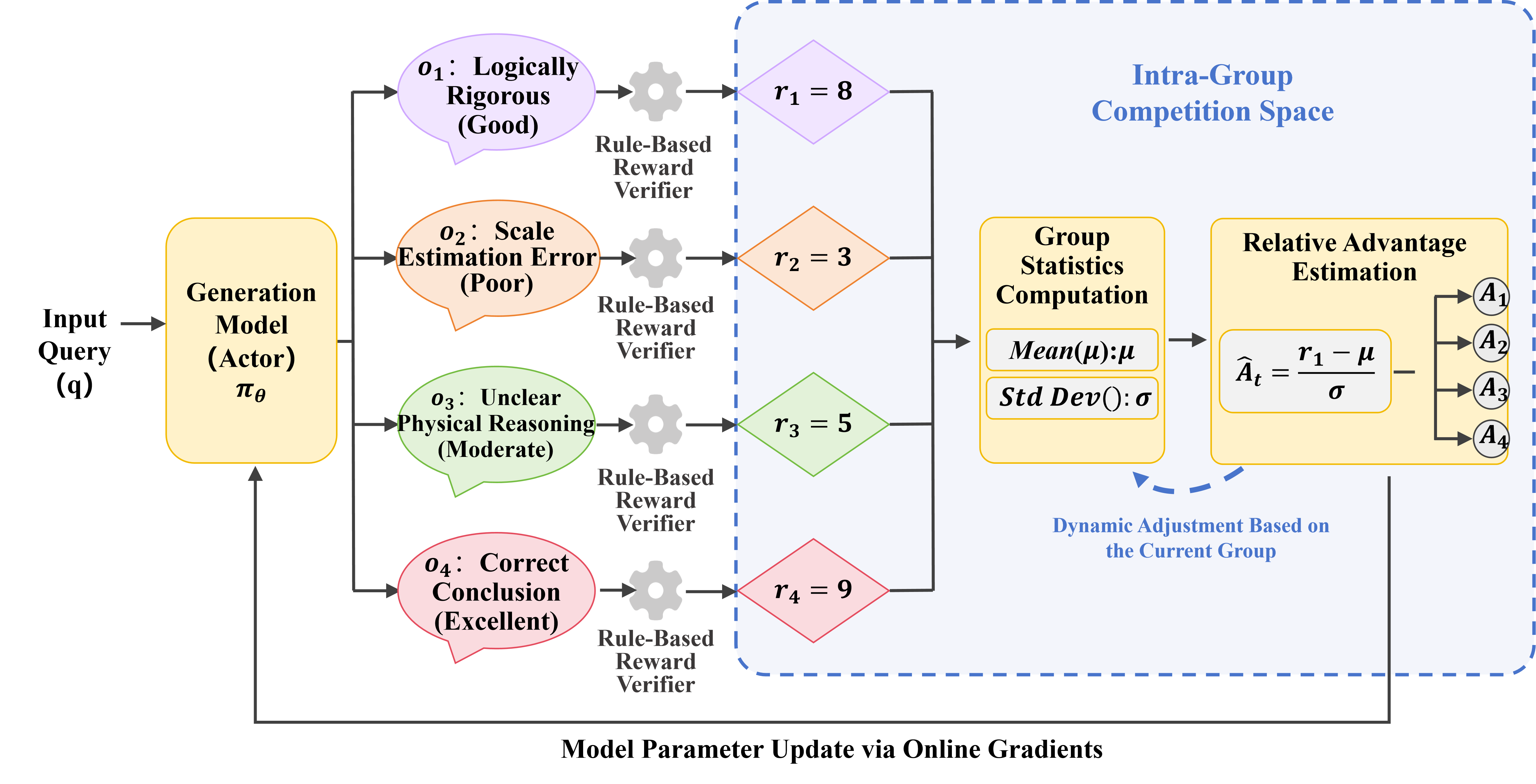}
    \caption{GRPO reinforcement learning process}
    \label{fig:fig04}
\end{figure}

Given that SAR interpretation chain-of-thought sequences are typically long and that candidate reasoning paths may exhibit substantial quality variations, this paper adopts GRPO for policy optimization. As shown in Figure \ref{fig:fig04}, the core mechanism of GRPO is to estimate relative advantages within a sampled group. This algorithm is based on the assumption that, for the same task instruction, multiple candidate outputs generated by the model through stochastic sampling may differ in logical rigor and physical consistency.

Instead of relying on a predefined absolute scoring criterion, GRPO determines the reward gradient by computing the relative position of the target sample $o_i$ within the current sampling group (G), thereby enabling iterative policy optimization. Mathematically, it uses the sample mean $\frac{1}{G} \sum r_j$ as a dynamic real-time estimate of the state value $V(s)$. The normalization formula is defined as follows:
\begin{equation}
	\hat{A}_i = \frac{r_i - \text{Avg}(\{r_1, \dots, r_G\})}{\text{Std}(\{r_1, \dots, r_G\})}
\end{equation}
This mechanism enables the model to compare the relative merits of multiple inference paths within the same task: inference patterns associated with high-reward samples are enhanced, while erroneous logic associated with low-reward samples is suppressed. Compared to offline preference optimization methods, GRPO does not rely on static preference pairs but obtains real-time feedback through online sampling and rule validation during training, making it more suitable for scenarios with complex physical constraints and long-chain inference in SAR interpretation.
To prevent policy drift or anomalous text generation when pursuing rule-based rewards, this paper further incorporates a Kullback Leibler (KL) divergence constraint into the GRPO objective, ensuring that the current policy $\pi_{\theta}$ does not deviate excessively from the reference policy. The optimization objective can be expressed as:
\begin{equation}
	\mathcal{J}_{GRPO}(\theta) = \frac{1}{G} \sum_{i=1}^G \left[ \mathcal{L}_{clip}(\theta) - \beta D_{KL}(\pi_{\theta} || \pi_{ref}) \right]
\end{equation}

Where $\mathcal{L}_{\mathrm{clip}}$ is the pruning strategy optimization term, and $\beta$ is the KL regularization coefficient. When the current strategy $\pi_{\theta}$ overfits to a specific heuristic reward during optimization (e.g., by generating redundant or anomalous text features to circumvent physical rule checks), the KL divergence term will significantly increase as a high-dimensional space constraint, thereby forcing the model state back to a reasonable semantic space through a powerful penalty term.

\subsubsection{Multi-task reward function} 

Reward function design is crucial in the reinforcement learning phase. Considering the multi-task nature of intelligent SAR image interpretation, this paper designs a unified reward system comprising sub-rewards such as format specification, target detection, target counting, category recognition, and ground feature classification.

\paragraph{Format reward function}

First, to ensure that the model output has a parsable reasoning structure, this paper designs a format reward $R_{\text{structure}}$ to constrain the model to simultaneously generate the thought process and the final answer:

\begin{equation}
R_{\text{structure}} = 
\begin{cases} 
1 & \text{including } \langle \text{think} \rangle \text{ and } \langle \text{answer} \rangle \\
0 & \text{else}
\end{cases}
\end{equation}

This reward can suppress the model's behavior of directly outputting the conclusion and skipping the reasoning process, and provide a clear text parsing boundary for subsequent task rewards.

\paragraph{Object Detection Reward Function}

Furthermore, for the object detection task, this paper designs an object detection reward, denoted as $R_{\mathrm{det}}$. This reward is calculated based on the intersection-over-union (IoU) between the predicted bounding box and the ground-truth bounding box, while also incorporating penalty terms for false positives and empty predictions. In this way, the reward provides explicit constraints on the model’s object localization capability and encourages more accurate and reliable detection results.

\begin{equation}
	R_{\text{det}} =
	\begin{cases}
		-\lambda_{\text{empty}} & N_g > 0 \land N_p = 0 \\
		1 & N_g = 0 \land N_p = 0 \\
		0 & N_g = 0 \land N_p > 0 \\
		\mathrm{MeanIoU} - \mathcal{L}_{fp} & \text{else}
	\end{cases}
\end{equation}

\paragraph{Target Counting and Classification Reward Function}

For the target counting and category recognition tasks, the model ultimately outputs a natural language answer in the `<answer>` tag, such as "The number of the target in the image: 3, the target category is aircraft". Therefore, it is necessary to first parse the predicted number of targets and the target category from the generated text. Let the predicted result parsed from the model output be $(\hat{n}, \hat{c})$, where $\hat{n}$ represents the predicted number of targets and $\hat{c}$ represents the predicted target category; the corresponding ground truth labels are denoted as (n, c), where n and c represent the actual number of targets and the target category, respectively. To simultaneously measure the model's performance in both "count prediction" and "category recognition" dimensions, this paper defines the reward for count and category matching as:

\begin{equation}
	R_{\text{num+type}}
	=
	\frac{1}{2}\,\mathbf{1}[\hat{n}=n]
	+
	\frac{1}{2}\,\mathbf{1}[\hat{c}=c]
\end{equation}

Here, $\mathbf{1}[\cdot]$ is an indicator function, which takes the value 1 when the condition is true and 0 otherwise. This reward can also be written in the following segmented form:

\begin{equation}
	R_{\text{num+type}}
	=
	\begin{cases}
		1 & \hat{n}=n \ \text{and}\ \hat{c}=c\\
		0.5 & \hat{n}=n \ \text{and}\ \hat{c}\neq c\\
		0.5 & \hat{n}\neq n \ \text{and}\ \hat{c}=c\\
		0 & \hat{n}\neq n \ \text{and}\ \hat{c}\neq c
	\end{cases}
\end{equation}

This design breaks down the task into two equally important sub-tasks: quantity prediction and category recognition. Compared to a binary reward of "1 for perfect accuracy and 0 for no accuracy" , this decomposed reward provides partial feedback even when the model only correctly predicts one dimension. This helps alleviate the sparse reward problem, improves training stability, and facilitates the model's gradual transition from coarse-grained accuracy to fine-grained perfect accuracy.

\paragraph{Ground feature classification task reward function}

For land-cover classification or region-level understanding tasks, this paper designs a land-cover reward, denoted as $R_{\mathrm{terrain}}$. Since land-cover interpretation may involve different task forms, such as single-label classification, multi-label prediction, and regional proportion estimation, a general land-cover reward framework is constructed to uniformly model these heterogeneous tasks. Let the model prediction be $\hat{y}$ and the ground-truth label be $y$. The predicted and ground-truth results are extracted using task-specific parsing functions. According to the corresponding task type, the reward function is defined as follows.

For the single-label classification task, the model is required to predict the dominant land-cover category in the image, such as water bodies or vegetation. Let the predicted category be $\hat{c}$ and the ground-truth category be $c$. The reward is then defined as:
\begin{equation}
	R_{\text{terrain}} = \mathbf{1}[\hat{c} = c]
\end{equation}
Where $\mathbf{1}[\cdot]$ is an indicator function, which takes a value of 1 when the prediction matches the true category, and 0 otherwise. This reward format is suitable for single-label classification tasks, emphasizing the model's accurate identification ability of dominant ground object semantics.

Multi-label classification task: In complex scenes, an image may simultaneously contain multiple land cover categories, such as $\{\text{water}, \text{road}, \text{building}\}$. Let the set of predicted categories be $\hat{C}$, and the set of true categories be $C$. Then, the reward is defined using set similarity (Jaccard coefficient): 
\begin{equation}
	R_{\text{terrain}} = \frac{|\hat{C} \cap C|}{|\hat{C} \cup C|}
\end{equation}
This reward can simultaneously consider both false negatives and false positives, and is more suitable for multi-label scenarios than simple accuracy.

For the spatial region classification task, where the input contains explicit spatial structures, the model is required to predict the land-cover category for each region independently. Let the predicted category of the (i)-th region be $\hat{c}_i$, and the corresponding ground-truth category be $c_i$, where $i \in \{1,2,3,4\}$. The reward is then defined as:
\begin{equation}
	R_{\text{terrain}} = \frac{1}{4} \sum_{i=1}^{4} \mathbf{1}[\hat{c}_i = c_i]
\end{equation}
This reward essentially performs an independent classification evaluation for each spatial sub-region and computes the average score to measure the model’s ability to capture spatial distribution information.

Land cover category proportion prediction task: This task requires estimating the proportion of different land cover features in an image (e.g., the area proportion of water bodies, vegetation, roads, and buildings). Let the predicted proportion be $\hat{p}_i$, and the true proportion be $p_i$, where $i \in \{1,2,3,4\}$. First, the mean absolute error (MAE) is calculated, and a continuous reward is defined based on the MAE:
\begin{equation}
	R_{\text{terrain}} = \max\left(0, 1 - \frac{\mathrm{MAE}}{\tau} \right)
\end{equation}
Where $\tau$ is the error tolerance threshold. This reward provides continuous feedback for regional land cover judgment, enabling the model to obtain a smoother optimization signal in complex land cover scenarios.

Finally, this paper weights and merges the above sub-rewards to obtain a unified reward function for multiple tasks:
\begin{equation}
	R = \alpha R_{\text{format}} + \beta  R_{\text{det}} + \gamma R_{\text{num+type}} + \delta R_{\text{terrain}}
\end{equation}


\section{Experiment}
\subsection{Experimental setup}
In terms of experimental configuration, this study constructed a refined hyperparameter system for the reinforcement learning process of FUSAR-R1. For the training strategy, the GRPO algorithm adopted a micro-learning rate of $\eta=1\times10^{-6}$ and a Warmup ratio of 0.01 to ensure robust evolution of the policy space during the initial exploration phase. For the long-range logical derivation involved in SAR image interpretation tasks, the system set the maximum generation length to 5500 tokens and used gradient accumulation technology to achieve deep optimization of long-sequence thought chains. In the sampling and reward mechanism, each training round samples $G=4$ candidate samples for the same instruction. A high sampling temperature (1.0) is maintained during the training phase to enhance the exploration space, while the temperature is reduced to 0 during the inference phase to ensure the determinism of the interpretation conclusion. Simultaneously, a KL divergence coefficient of $\beta=0.001$, combined with a multi-dimensional reward function with a weight range of $[0.8, 1.0]$, effectively balances format regularity and physical perception accuracy. At the system architecture level, this study maximizes graphics processing unit (GPU) memory utilization by adopting the DeepSpeed ZeRO-2 optimizer and bfloat16 numerical precision. In addition, an asymmetric computational resource allocation strategy is designed to improve training efficiency under large-scale model settings. With the collaborative operation of an eight-A100 GPU cluster, real-time sampling and parameter updates can be efficiently performed, ensuring stable convergence and efficient iteration of the SAR reasoning model during large-scale reinforcement learning optimization.

\begin{table}[t]
\centering
\caption{Key Hyperparameter Settings for GRPO Training and Inference}
\label{tab:grpo_config}

\footnotesize
\renewcommand{\arraystretch}{1.15}
\setlength{\tabcolsep}{3pt}

\begin{tabularx}{\columnwidth}{
@{}
>{\centering\arraybackslash}p{1.5cm}
>{\centering\arraybackslash}X
>{\centering\arraybackslash}p{1.5cm}
@{}
}
\toprule
\textbf{Category}& 
\centering\textbf{Parameter} &
\textbf{Value} \\
\midrule

\multirow{5}{*}{\makecell{GRPO\\Training}}
& Learning rate $\eta$ & $1\times10^{-6}$ \\
& Training epochs & 1 \\
& Training batch size & 1 \\
& Gradient accumulation & 2 \\
& Warmup ratio & 0.01 \\
\midrule

\multirow{4}{*}{\makecell{Generation\\Strategy}}
& Maximum generation length & 5500 \\
& Number of generated samples & 4 \\
& Training temperature & 1.0 \\
& Inference temperature & 0 \\
\midrule

\multirow{3}{*}{\makecell{Reward\\Function}}
& Reward functions &
\begin{tabular}[t]{@{}c@{}}
Format and\\[-2pt]
task rewards
\end{tabular}
\\[-1pt]

& Reward weights & $[0.8,\,1.0]$ \\
& KL coefficient $\beta$ & 0.001 \\
\midrule

\multirow{4}{*}{\makecell{System\\Configuration}}
& Precision & bfloat16 \\
& DeepSpeed stage & ZeRO-2 \\
& Data loading workers & 4 \\
& Maximum new tokens & 4000 \\
\midrule

\multirow{3}{*}{\makecell{Distributed\\Setting}}
& Rollout GPUs & 2 GPUs \\
& RLHF GPUs & 6 GPUs \\
& Inference GPUs & 8 GPUs \\
\bottomrule
\end{tabularx}

\end{table}

\subsection{Experimental results}
To comprehensively and objectively evaluate the actual performance of FUSAR-R1 at different interpretation levels, this study designed six evaluation tasks covering target perception, semantic analysis, and quantitative reasoning (as shown in Table \ref{tab:task_definition}).

\begin{table}[t]
\centering
\caption{FUSAR-R1 Inference Capability Evaluation Tasks}
\label{tab:task_definition}

\scriptsize
\renewcommand{\arraystretch}{1.12}
\setlength{\tabcolsep}{2pt}

\begin{tabularx}{\columnwidth}{
@{}
>{\centering\arraybackslash}p{0.8cm}
>{\centering\arraybackslash}p{1.8cm}
>{\RaggedRight\arraybackslash}X
>{\centering\arraybackslash}p{0.8cm}
@{}
}
\toprule

\textbf{Task ID} &
\textbf{Task Type} &
\centering\arraybackslash
\textbf{Expected Output and Evaluation Dimension} &
\textbf{Samples} \\

\midrule

Task 1 &
\makecell[c]{Object\\Detection} &
\texttt{[\{label: aircraft, box: [...]\}, ...]}
\newline
\textit{Evaluation:} Spatial localization capability &
9000 \\

\addlinespace[2pt]

Task 2 &
\makecell[c]{Object\\Classification\\and Counting} &
Object number: 3; category: aircraft
\newline
\textit{Evaluation:} Object classification and counting accuracy &
9000 \\

\addlinespace[2pt]

Task 3 &
\makecell[c]{Main Land\\Cover Category} &
Main land cover category: [vegetation]
\newline
\textit{Evaluation:} Global semantic perception &
600 \\

\addlinespace[2pt]

Task 4 &
\makecell[c]{Comprehensive\\Land Cover\\Recognition} &
Land cover categories: [vegetation, building, water]
\newline
\textit{Evaluation:} Recognition completeness in complex scenes &
600 \\

\addlinespace[2pt]

Task 5 &
\makecell[c]{Regional Land\\Cover Analysis} &
Upper left, upper right, lower left, and lower right:
[vegetation, building, ...]
\newline
\textit{Evaluation:} Spatial semantic reasoning &
600 \\

\addlinespace[2pt]

Task 6 &
\makecell[c]{Land Cover\\Proportion\\Estimation} &
Water: 10\%; building: 90\%
\newline
\textit{Evaluation:} Quantitative land cover proportion estimation &
600 \\

\bottomrule
\end{tabularx}
\end{table}

Tasks 1 and 2 focus on basic target-level interpretation, examining not only the model's ability to accurately classify and count SAR targets but also verifying the model's robustness to positioning under complex commands through formatted output requirements. Tasks 3 through 5 extend the evaluation scope to areal features, examining the model's completeness of perceiving the macroscopic semantic background of SAR images through the extraction of feature categories and spatial distribution descriptions at different granularities.

The multi-dimensional task settings not only avoid the limitations of overly simplistic evaluation metrics but also accurately reflect the consistency and reliability of the model's logical reasoning when facing diverse and practical SAR interpretation needs. To delve into the sources of the FUSAR-R1 model's reasoning ability and verify the contribution of each training stage to performance improvement, this paper adopts a progressive experimental verification scheme. First, using the object detection task as the core evaluation benchmark, detailed ablation experiments are conducted to analyze the impact of cold start and reinforcement learning strategies on the inference model's capabilities. Subsequently, using the optimal model configuration verified through ablation experiments, comparative experiments are performed with mainstream multimodal models on the remaining five downstream tasks to comprehensively evaluate the model's multi-task generalization performance.

\subsubsection{Target detection task}

This task requires the model to output bounding box coordinates and category information in a strict JSON format, making it one of the most representative tasks in intelligent SAR image interpretation. Table \ref{tab:ablation_results} shows the impact of different modules on model performance.

\begin{table}[t]
	\centering
    \scriptsize
	\caption{Ablation Study Results of FUSAR-R1}
	\label{tab:ablation_results}
	
	\renewcommand{\arraystretch}{1.1}
	\setlength{\tabcolsep}{3pt}
	
	\begin{tabular}{@{} c c c c c c @{}}
		\toprule
		\textbf{No.} & 
		\textbf{CoT Cold Start} & 
		\textbf{Reinforcement Learning} & 
		\textbf{Precision} & 
		\textbf{Recall} & 
		\textbf{F1} \\ 
		\midrule
		
		1 & \ding{55} & \ding{55} & 0.639 & 0.449 & 0.528 \\
        
		2 & \ding{51} & \ding{55} & 0.723 & 0.513 & 0.600 \\
		
		3 & \ding{55} & GRPO & 0.370 & 0.134 & 0.197 \\
		
		4 & \ding{55} & CISPO & 0.361 & 0.131 & 0.192 \\
		
		5 & \ding{52} & CISPO & 0.798 & 0.570 & 0.665 \\
		
		6 & \ding{52} & GRPO & \textbf{0.807} & \textbf{0.568} & \textbf{0.676} \\
		
		\bottomrule
	\end{tabular}
\end{table}

Experiment 6 achieved the best performance across all metrics, with precision, recall, and F1 score reaching 0.807, 0.568, and 0.676, respectively. This result significantly outperforms other configuration combinations, demonstrating the effectiveness of combining cold-start training with GRPO. Therefore, this study uses it as the default training configuration for FUSAR-R1, clarifying its relevance to the study's setup.

A comparison between Experiment 1 and Experiments 3 and 4 shows that directly applying reinforcement learning without cold-start training leads to performance collapse. This result indicates that, for SAR image interpretation tasks, the model must first acquire basic physical reasoning logic and standardized output specifications through the cold-start phase. Otherwise, an effective reward feedback mechanism cannot be established within the large-scale search space of reinforcement learning. A comparison between Experiment 2 and Experiments 5 and 6 further reveals a second-stage improvement in model performance. By introducing reinforcement learning on the basis of cold-start training, the model is able to autonomously explore reasoning paths and optimize interpretation strategies. Through continuous error correction and reinforcement of chain-of-thought logic, the model achieves significant improvements in localization accuracy and target recognition capability.

In summary, the ablation experiment results fully validate the effectiveness of the inference model construction paradigm of "cold start injection logic followed by reinforcement learning optimization strategy" proposed in this paper.

\subsubsection{Comparison with other advanced large models}

To further evaluate the generalization performance of FUSAR-R1 in general remote sensing interpretation scenarios, this study selects five representative open-source multimodal large models for comparison, including InternVL 3.5, LLaVA 1.5, and Qwen2.5-VL. Tables \ref{tab:model_comparison} present the quantitative results of each model across different downstream tasks.

\begin{table}[t]
	\centering
	\caption{Performance Comparison With Mainstream Multimodal Large Models}
	\label{tab:model_comparison}
	\scriptsize
	\renewcommand{\arraystretch}{1.1}
	\setlength{\tabcolsep}{2pt}
	
	\begin{tabular}{p{2.25cm} p{1.15cm} p{1.05cm} p{1.05cm} p{1.05cm} p{0.9cm}}
		\toprule
		\textbf{Model} & 
		\textbf{Object Counting (Acc.) (\%)} & 
		\textbf{Main Land-cover Category (Acc.) (\%)} & 
		\textbf{All Land-cover Categories (Acc.) (\%)} & 
		\textbf{Regional Land-cover Category (Acc.) (\%)} & 
		\textbf{Land-cover Proportion (MAE)} \\ 
		\midrule
		
		InternVL3\_5\_4B   & 45.45 & 44.23 & 75.81 & 6.55  & 39.58 \\
		
		InternVL3\_5\_8B   & 41.41 & 8.19  & 5.60  & 3.27  & 90.09 \\
		
		Llava\_1\_5\_7b     & 40.91 & 9.83  & 62.15 & 12.29 & 57.15 \\
		
		Qwen2\_5\_VL\_3B   & 35.35 & 8.19  & 35.10 & 9.42  & 40.75 \\
		
		Qwen2\_5\_VL\_7B   & 34.85 & 36.06 & 65.30 & 13.11 & 56.23 \\ 
		
		\midrule
        \rowcolor{gray!20}
		\textbf{FUSAR-R1} & 
		\textbf{67.33} & 
		\textbf{85.24} & 
		\textbf{93.44} & 
		\textbf{64.75} & 
		\textbf{7.67} \\ 
		
		\bottomrule
	\end{tabular}
\end{table}

Experimental results demonstrate that FUSAR-R1 exhibits significant advantages in multiple interpretation tasks. In the tasks of identifying major land cover categories and all land cover categories, its accuracy reaches 85.24\% and 93.44\%, respectively, far exceeding the corresponding metrics of the suboptimal model InternVL3.5-4B. In the challenging tasks of target counting and regional land cover category identification, FUSAR-R1 achieves accuracies of 67.33\% and 64.75\%, respectively, while the regional identification accuracy of the comparison models is generally below 15\%. Furthermore, in terms of quantitative inference capability, FUSAR-R1's mean absolute error is only 7.67, representing a substantial improvement over general-purpose models. The data comparison also reflects that model size is not the determining factor for performance; with a moderate parameter size, FUSAR-R1 comprehensively surpasses the participating ultra-large-scale general-purpose models in all interpretation metrics.

\subsubsection{Analysis of key factors affecting reasoning}

To reveal the underlying logical support role of different dimensions of physical priors in the FUSAR-R1 inference chain, this section conducts a two-way ablation experiment. By comparing the experimental results of "single-item injection" and "single-item removal" (see Table VI), the following conclusions are drawn: The experimental results show that under the condition of only including the basic visual modality (labeled as SAR), the detection performance of the model is limited (F1 score is only 0.324). However, when image physical size information is introduced alone, the model performance shows explosive growth, with the F1 score increasing to 0.583. This indicates that in SAR image interpretation, spatial scale constraints are the core premise for the model to establish target physical size perception, anchoring visual features from simple pixel distribution to a real physical measurement space. In contrast, single priors such as satellite category or band have relatively small marginal gains on model performance when scale benchmarks are lacking.

\begin{table}[t]
\centering
\scriptsize
\caption{Ablation Results of Different Physical Priors on FUSAR-R1 Performance}
\label{tab:ablation_study_fusarr1}

\renewcommand{\arraystretch}{1.1}
\setlength{\tabcolsep}{3pt}

\begin{tabular}{p{6.0cm}ccc}
\toprule
\textbf{Configuration / Experimental Setting} & 
\textbf{P} & 
\textbf{R} & 
\textbf{F1} \\
\midrule

\rowcolor{gray!20} 
\textbf{Full Baseline (Resolution, Size, Band, Location, Satellite, Modality)} 
& \textbf{0.807} & \textbf{0.568} & \textbf{0.676} \\

\midrule

\textit{Single Information Injection} & & & \\

\quad Only Image Size & 0.728 & 0.542 & 0.583 \\

\quad Only Resolution & 0.374 & 0.287 & 0.325 \\

\quad Only Band & 0.381 & 0.278 & 0.321 \\

\quad Only Location & 0.243 & 0.154 & 0.188 \\

\quad Only Satellite & 0.318 & 0.231 & 0.268 \\

\quad Only Modality Information (SAR) & 0.382 & 0.282 & 0.324 \\

\midrule

\textit{Single Information Removal} & & & \\

\quad Only Removing Image Size & 0.395 & 0.276 & 0.325 \\

\quad Only Removing Resolution & 0.795 & 0.573 & 0.666 \\

\quad Only Removing Band & 0.788 & 0.551 & 0.669 \\

\quad Only Removing Location & 0.785 & 0.545 & 0.659 \\

\quad Only Removing Satellite & 0.803 & 0.566 & 0.664 \\

\quad Only Removing Modality Information (SAR) & 0.802 & 0.559 & 0.675 \\

\bottomrule
\end{tabular}

\end{table}

Ablation experiments demonstrate that physical scale and image resolution provide essential constraints for FUSAR-R1 to achieve accurate interpretation, establishing the fundamental basis for detection accuracy. Meanwhile, auxiliary information, such as satellite type and geographic location, further enhances model performance through multi-dimensional contextual guidance, enabling FUSAR-R1 to achieve optimal interpretation results.

\subsubsection{Visualization Analysis}
\paragraph{Evaluation of training convergence}

\begin{figure}
    \centering
    \includegraphics[width=1\linewidth]{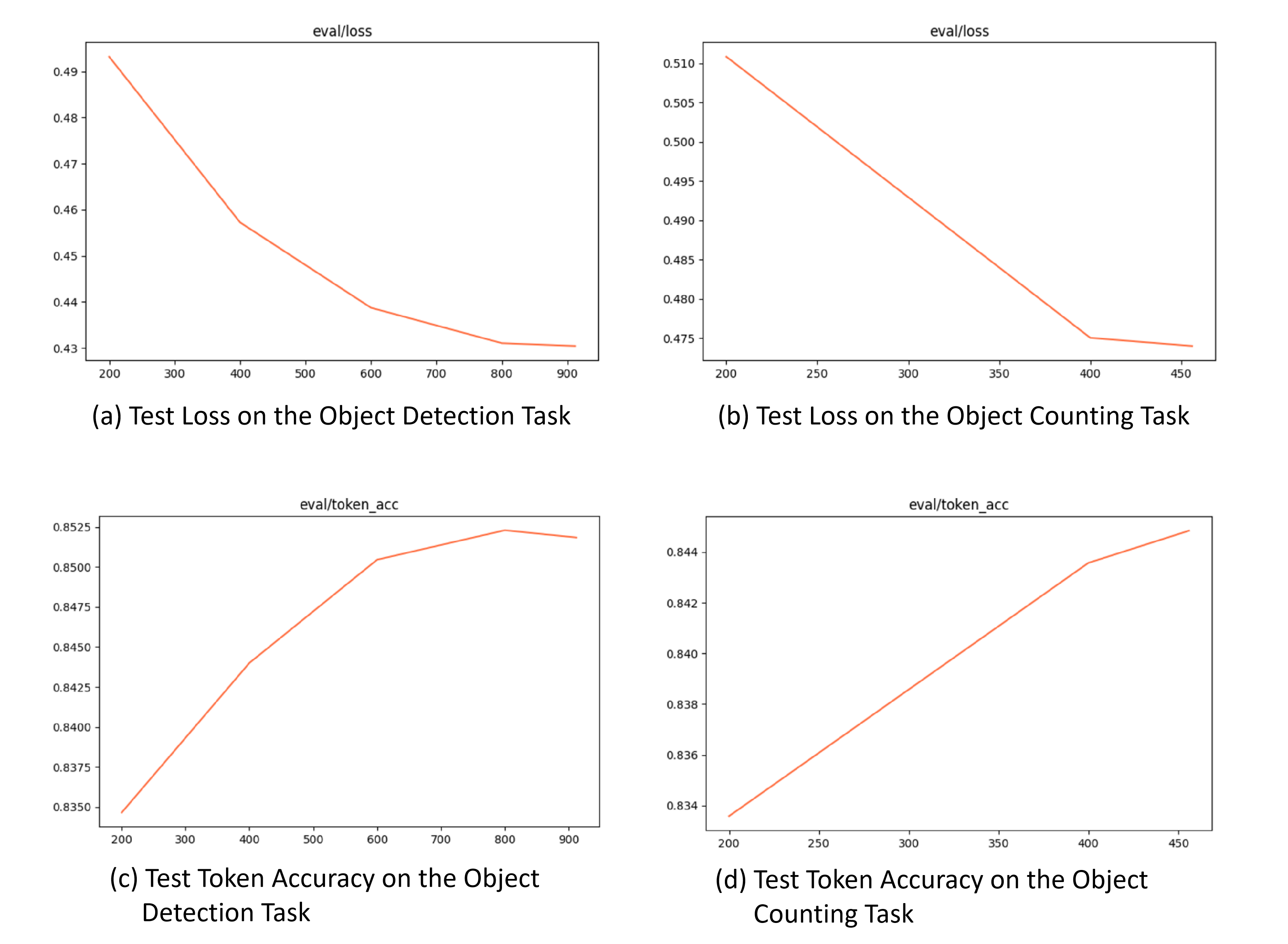}
    \caption{Performance metrics of the FUSAR-R1 training process}
    \label{fig:Performance_metrics}
\end{figure}

This section quantitatively analyzes the convergence performance of FUSAR-R1 during the training phase and further verifies its effectiveness in handling complex SAR image interpretation tasks by combining the reasoning processes of representative task scenarios. Figure \ref{fig:Performance_metrics} presents the training curves of the model for target detection and target counting tasks. As the training steps increase, the loss values of both tasks on the test set exhibit a steady and significant decreasing trend. The target detection task reaches a stable plateau after approximately 800 steps, indicating that the model has effectively learned the mapping relationship between SAR imagery and the underlying reasoning process.

\paragraph{Typical scenario reasoning process}

\begin{figure}
    \centering
    \includegraphics[width=1\linewidth]{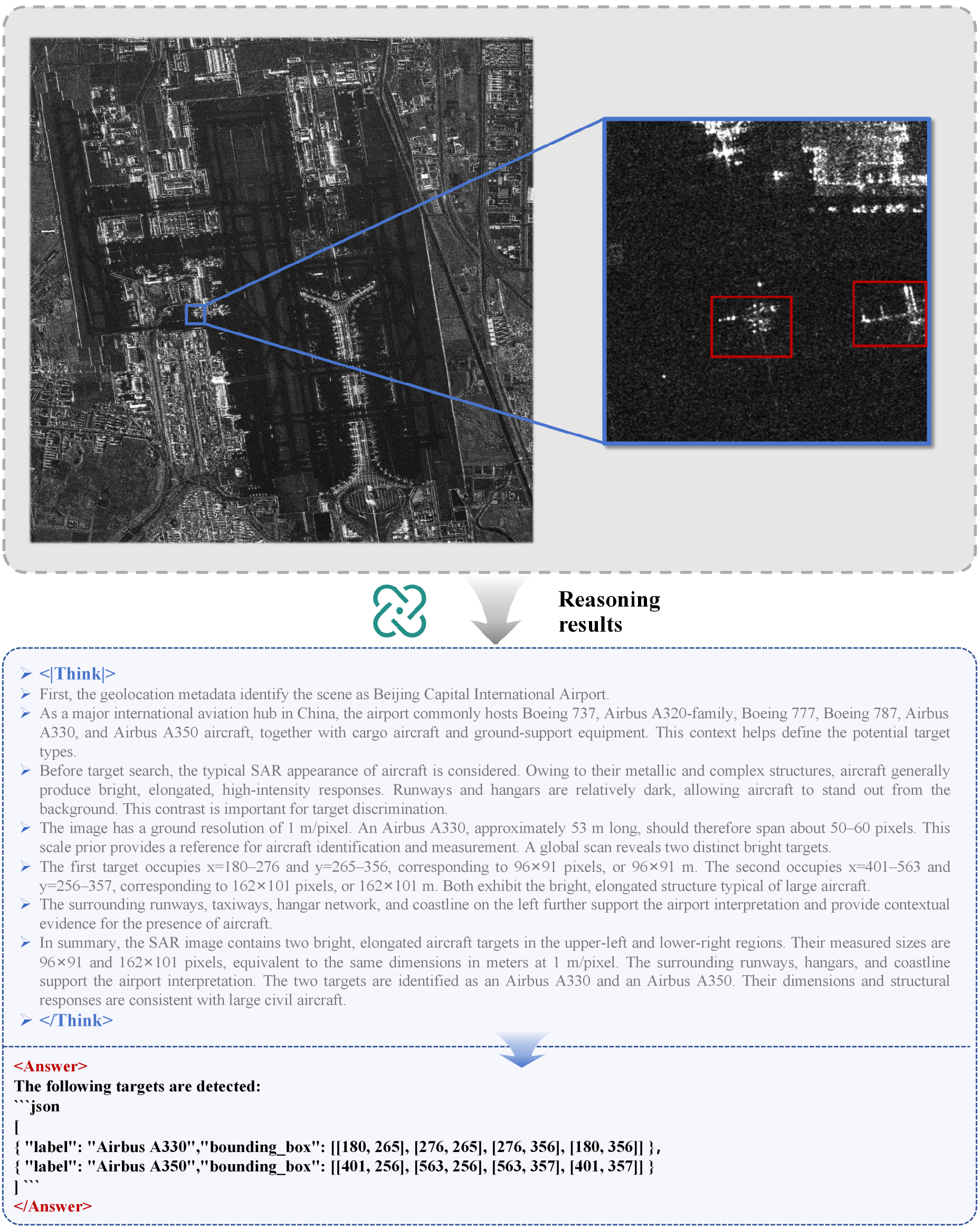}
    \caption{Airport aircraft reasoning examples}
    \label{fig:fig06}
\end{figure}

\begin{figure}
    \centering
    \includegraphics[width=1\linewidth]{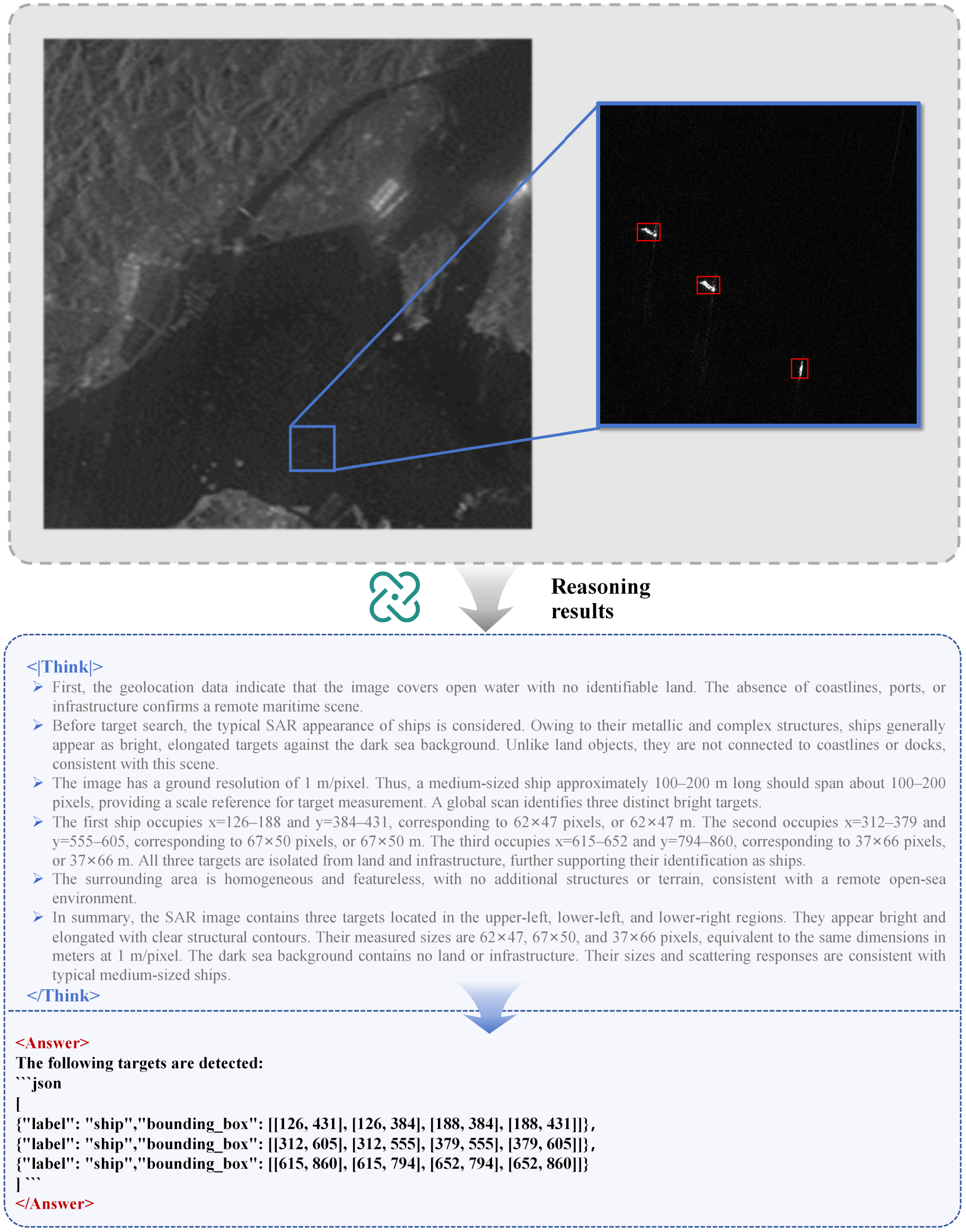}
    \caption{Reasoning examples for surface ships}
    \label{fig:fig07_ship}
\end{figure}

To further quantify and verify the logical rigor of FUSAR-R1 in complex real-world interpretation tasks, this section selects two typical scenarios: a civil aviation hub (Figure \ref{fig:fig06}) and open sea (Figure \ref{fig:fig07_ship}), and conducts a retrospective analysis of their entire reasoning process.

Aircraft target detection and recognition in the airport hub scenario: As shown in Figure \ref{fig:fig06}, in the SAR image interpretation task for Beijing Capital International Airport, FUSAR-R1 demonstrates its logical reasoning ability from macroscopic background to microscopic features.
The model first extracts the input latitude and longitude priors, identifying the current scene as a large international aviation hub. Based on this background knowledge, the model pre-sets Boeing and Airbus series commercial airliners as potential target candidates in its thought process. Then, combining an imaging resolution of 1 meter per pixel, the model accurately maps the pixel span (approximately 96 × 91 pixels) of bright scattering clusters in the image to their physical size, effectively eliminating interference from small support vehicles or runway clutter. Finally, by recognizing the bright, slender structure of the target with typical metallic scattering characteristics, the model accurately determines the presence of two targets in the scene: an Airbus A330 and an Airbus A350, and outputs the precise detection box coordinates in standard JSON format.

Ship Target Detection and Recognition in Open Sea Scenarios: Figure \ref{fig:fig07_ship} illustrates the inference performance of the model in scenarios without prominent geographic references. The model first analyzes the overall scattering characteristics of the image to determine the absence of coastlines, ports, and artificial infrastructures, thereby identifying the scene as a “clean sea surface background”. During the subsequent search process, the model detects three isolated high-intensity scattering targets. According to the inference process, the model estimates their physical lengths in real time based on the 1 m spatial resolution, obtaining values of 62 m, 67 m, and 37 m, respectively. Despite the presence of random speckle noise over the sea surface, the model effectively reduces false alarms caused by sea clutter by evaluating the consistency between target scattering intensity and the structural characteristics of medium-sized ships. Ultimately, the model identifies all three targets as ships, demonstrating the robustness and reliability of its decision-making process in low signal-to-noise ratio environments.

The above examples demonstrate that the core advantage of FUSAR-R1 lies in its deep integration of visual perception and interpretive reasoning. Rather than relying solely on pixel-level statistical features, the model incorporates geographic priors and spatial resolution information to perform closed-loop logical reasoning, thereby significantly improving the reliability and interpretability of SAR image interpretation results in complex and large-scale scenarios.

\section{conclusion}

This paper focuses on the core challenges of SAR cognitive intelligent models, including the lack of transparency in the interpretation process and the limited credibility of model decisions. To bridge the gap from semantic understanding to logical reasoning, this paper proposes FUSAR-R1, a large-scale reasoning model based on reinforcement learning and chain-of-thought mechanisms, establishing an interpretable and highly reliable reasoning framework for intelligent SAR interpretation. Experimental results demonstrate that, by developing a training paradigm that integrates chain-of-thought cold start and reinforcement learning optimization, FUSAR-R1 achieves substantial improvements in downstream task performance, with accuracy and F1-score increasing by 18\% and 15\%, respectively, compared with the baseline model. Furthermore, in comparison with open-source mainstream multimodal large models of general-purpose, FUSAR-R1 achieves an average performance improvement of over 30\% on key tasks, including target counting and land-cover classification. Furthermore, the mean absolute error (MAE) metric, which measures the predictive power of land cover category proportions, showed a significant reduction in error of over 31.91, validating the model's superior reasoning and quantitative analysis capabilities in complex remote sensing scenarios.

This research represents a crucial leap in SAR intelligent interpretation from "perception-cognition" to "reasoning". By integrating the thought chain mechanism with reinforcement learning, it not only solves the interpretability challenge of deep learning models but also enhances the reliability of interpretation in complex scenarios, laying a vital foundation for the practical application of large-scale SAR intelligent interpretation models.

\bibliographystyle{unsrt}
\bibliography{main}

\end{document}